\documentclass[10pt,twocolumn,letterpaper]{article}

\usepackage{wacv}
\usepackage{times}
\usepackage{epsfig}
\usepackage{graphicx}
\usepackage{amsmath}
\usepackage{amssymb}
\usepackage{multirow}
\usepackage{nopageno}
\usepackage{balance}
\usepackage[dvipsnames]{xcolor}
\newcommand{\ACCVmodifica}[1]{\textcolor{black}{#1}}
\newcommand{\WACVmodifica}[1]{\textcolor{black}{#1}}
\newcommand{\JC}[1]{}
\newcommand{\FM}[1]{}
\newcommand{\VM}[1]{}

\def\wacvPaperID{870} % Enter the WACV Paper ID here

\wacvfinalcopy % *** Uncomment this line for the final submission

\ifwacvfinal
\def\assignedStartPage{0000} % *** Enter the assigned starting page number (instead of 9876)
\fi

\ifwacvfinal
\usepackage[breaklinks=true,bookmarks=false]{hyperref}
\else
\usepackage[pagebackref=true,breaklinks=true,colorlinks,bookmarks=false]{hyperref}
\fi

\ifwacvfinal
\else
\fi

\begin{document}

%%%%%%%%% TITLE
\title{Transductive Zero-Shot Learning by Decoupled Feature Generation}

\author{Federico Marmoreo$^{1,2}$, Jacopo Cavazza$^1$ and Vittorio Murino$^{1,3,4}$\\
$^1$ Pattern Analysis and Computer Vision, Istituto Italiano di Tecnologia, Italy   \\
%$^2$ Universit\`{a} degli Studi di Genova, Italy \\
$^2$ University of Genova, Italy \\
$^3$ Huawei Technologies Ltd., Ireland Research Center, Ireland \\
$^4$ Department of Computer Science, University of Verona, Italy \\
{\tt\small \{federico.marmoreo,jacopo.cavazza,vittorio.murino\}@iit.it}
% \and
% Jacopo Cavazza\\
% \and
% Vittorio Murino\

}

\maketitle
\thispagestyle{empty}

\begin{abstract}
In this paper, we address zero-shot learning (ZSL), the problem of recognizing categories for which no labeled visual data are available during training.
We focus on the transductive setting, in which unlabelled visual data from unseen classes is available. State-of-the-art paradigms in ZSL typically exploit generative adversarial networks to synthesize visual features from semantic attributes. 
We posit that the main limitation of these approaches is to adopt a single model to face two problems: 1) generating realistic visual features, and 2) translating semantic attributes into visual cues. Differently, we propose to \emph{decouple} such tasks, solving them separately. In particular, we train an unconditional generator to solely capture the complexity of the distribution of visual data and we subsequently pair it with a conditional generator devoted to enrich the prior knowledge of the data distribution with the semantic content of the class embeddings. We present a detailed ablation study to dissect the effect of our proposed \emph{decoupling} approach, while demonstrating its superiority over the related state-of-the-art.
\end{abstract}

\section{Introduction}

The field of visual object recognition has seen a significant progress in recent years, mainly because of the availability of large-scale \emph{annotated} datasets. However, labelling data is not only difficult and costly but is also prone to errors since requiring human intervention. Furthermore, annotating a big corpus of data for each of the categories to be recognized in a balanced way is simply unfeasible due the well known long-tail distribution problem \cite{salakhutdinov2011learning}.

As a promising solution to the aforementioned issues, \emph{zero-shot learning} (ZSL) algorithms tackle the problem of

recognizing novel categories, even if a classifier has not been directly trained on them \cite{larochelle2008zero,lampert2009learning}.
In a fully supervised paradigm, each single category is assumed to be (evenly) represented by a set of annotated visual data (images). ZSL methods instead, allow this assumption to hold only for a restricted set of
\emph{seen} categories. 
The goal is then to recognize a disjoint set of target \emph{unseen} categories, when, in the transductive setting, only unlabelled visual data is available (differently from the inductive setting, in which no visual data from unseen classes are available).
To transfer from seen to unseen categories, auxiliary information is typically adopted in the form of either manually-defined attributes \cite{lampert2009learning} or distributed word embeddings \cite{xian2018tPAMI}. 
When transferring from the seen to the unseen classes, the main challenge is handling such category shift: in this paper, we evaluate this in the \emph{generalized} ZSL (GZSL) setup in which a method is evaluated on both seen and unseen classes, requiring to learn the latter without forgetting the former ones. In fact, given the separation of the data into labelled seen and unlabelled unseen instances, supervised training can be done for seen classes only, resulting in an unbalanced performance in GZSL. 
To address this issue, various recent works proposed to augment the unseen-class data with synthetic labeled data \cite{mishra2018generative,Arora2018GeneralizedZL,felix2018multi,huang2019generative,Li:CVPR19,Xian_2019_CVPR,Schonfeld_2019_CVPR,Zhu:ICCV19,Xian_2018_CVPR,Sariyildiz2019GradientMG,Gao2020ZeroVAEGANGU}.

In this paper, we address the transductive GZSL problem, by introducing a novel, more effective, feature synthesis method able to balance the training process.
In details, our approach builds upon 
the possibility of mimicking the human brain in hallucinating a mental imagery of a certain unknown category while reading a textual description of it. As pursued by a number of recent works \cite{mishra2018generative,Arora2018GeneralizedZL,felix2018multi,huang2019generative,Li:CVPR19,Xian_2019_CVPR,Schonfeld_2019_CVPR,Zhu:ICCV19,Xian_2018_CVPR,Sariyildiz2019GradientMG,Gao2020ZeroVAEGANGU}, conditional feature generation is adopted for this purpose. Specifically, images from the seen classes are fed into a backbone ResNet-101 network which pre-computes a set of real visual features \cite{xian2018tPAMI}. Subsequently, through a Generative Adversarial Network (GAN) \cite{Goodfellow2014GenerativeAN,Arjovsky2017WassersteinG,improvedWGAN}, the following min-max optimization game is solved: a generator network is asked to synthesize visual embeddings which should look real to a discriminator module. Since the generator network is conditionally dependent upon semantic embeddings, the trained model can be exploited to create synthetic features of the classes for which we lack visual labeled data. Afterwards, GZSL can be solved as simple classification problems through a softmax classifier trained on top of real features (from seen classes) and generated features (from the unseen ones). 

In the literature, several variants have been attempted to improve the pipeline under the architectural point of view with the aim to solving efficiently GZSL: 
using an attribute regression module \cite{huang2019generative},
easing the generator with a variational auto-encoder \cite{mishra2018generative,Arora2018GeneralizedZL,Xian_2019_CVPR,Schonfeld_2019_CVPR,Gao2020ZeroVAEGANGU}, adopting cycle-consistency \cite{felix2018multi}, designing intermediate latent embeddings \cite{Li:CVPR19} or employing features-to-feature translation methods \cite{mishra2018generative,Arora2018GeneralizedZL,felix2018multi,huang2019generative,Li:CVPR19,Xian_2019_CVPR,Schonfeld_2019_CVPR,Zhu:ICCV19,Xian_2018_CVPR,Sariyildiz2019GradientMG,Gao2020ZeroVAEGANGU}.
On the one hand, as commonly done in adversarial training, we need to generate synthetic
descriptors indistinguishable from a pool of pre-trained real features used as reference. On the other hand, synthetic 
features are required to translate the semantic information into visual patterns, which are discriminative for the seen and unseen classes to be recognized. 

We posit that resolving these two tasks within a single architecture is arguably difficult and we claim that this is the major limitation affecting the performance of the currently available feature generating schemes for GZSL. In fact, since adopting a single architecture for two tasks, one of them may be suboptimally solved with respect to the other, resulting in a poor modelling of either the visual or the semantic space.

Hence, differently to prior work, in this paper we propose to separately solve the two tasks, \emph{decoupling} the feature generation stage to better tackle transductive GZSL. First, we train an unconditional generative adversarial network with the purpose of synthesizing features visually similar to the real ones in order to properly model their distribution in an unsupervised manner. Since our generation is not conditioned on semantic embeddings, we are sure to specifically model the visual appearance of our feature representations. Second, we encapsulate such visual information into a \emph{structured prior}, which is used in tandem with a conditioning factor (here, the semantic embedding) to (conditionally) generate synthetic feature vectors. Because of the improved source of noise, we expect to enhance the semantic-to-visual translation as well, yielding visual descriptors with richer semantic content. The resulting architecture for decoupled feature generation is named DecGAN.

Since our DecGAN is decoupled into one unconditional and one conditional 
GAN-like branches, it is capable of exploiting the unlabeled visual data which are available in transductive GZSL.
In fact, while we can compare the generated seen features with the real ones, both conditionally and unconditionally (since we have access to labels), we cannot do it 
for the unseen ones. Unseen classes are, in fact, not supported by annotated visual data, hence the conditional discriminator cannot ``verify'' them. 
We deem that our proposed architecture contributes in addressing this problem by \emph{cross-connecting} the conditional branch with the unconditional one. In other words, we use the unconditional discriminator to evaluate the ``quality'' of the conditionally generated features. In this way, we decouple the feature generation, not only for the seen categories, but also for the unseen ones, resulting in a better modelling for both and improving the GZSL performance.

In summary, this work provides the following original contributions.
\begin{itemize} 
    \item We introduce the idea of \emph{decoupling} feature generation for transductive zero-shot learning by encapsulating visual patterns into a structured 
    prior, which is subsequently adopted to boost the semantically conditioned synthesis of visual features.
    \item We implement our idea through a novel architecture, termed DecGAN, which combines an unconditional and a conditional feature generation module, introducing a novel \emph{cross-connected branch} mechanism
    able to decoupling feature generation for both seen and unseen categories.
    \item Through an extensive ablation study, we analyze each single component of our architecture. As  compared to the transductive GZSL state-of-the-art,
    DecGAN outperforms it on CUB \cite{CUB} and SUN \cite{SUN} datasets (see Table \ref{tab:results_trans_ZSL_GZSL}, Section \ref{sec:exp}).
\end{itemize}
The rest of the paper is organised as follows. In Section \ref{sec:relwork}, we outline the most relevant related work. In Section \ref{sec:method}, we present the proposed approach, which is then experimentally validated in Section \ref{sec:exp}. The final conclusions are drawn in Section \ref{sec:conc}.

\section{Related Work}\label{sec:relwork}

In this Section, we will cover the most relevant related work in the field of transductive zero-shot learning. For more general details on other zero-shot paradigms, the reader can refer to \cite{xian2018tPAMI}.

\ACCVmodifica{Classical approaches in GZSL aim at learning a compatibility function between visual features and class embeddings, projecting them in a common space \cite{xian2018tPAMI}.}

In order to exploit unlabeled data for the unseen classes in the \ACCVmodifica{transductive GZSL},  \cite{Ye2017ZeroShotCW} proposes a procedure to perform label propagation \ACCVmodifica{\cite{Fujiwara2014EfficientLP}} 
as to simultaneously learn a representation for both seen and unseen classes. 
As shown in \cite{xian2018tPAMI}, this label propagation procedure can be extended to all the methods based on compatibility functions that map the visual features into the class embedding space \cite{ALE}. 
\ACCVmodifica{Due to the shift between seen and unseen classes, the projection may struggle when switching from the seen to the unseen domain. To this end, in \cite{Wan2019TransductiveZL,EDE_2018,Zhao2018DomainInvariantPL}, the projection function is improved, while \cite{Ye2019ProgressiveEN} tries to alleviate the issue using an ensemble of classifiers}. Differently, we perform synthetic feature generation to produce labeled visual features for the unseen classes.

Generative approaches for transductive \ACCVmodifica{GZSL} have been recently proposed \ACCVmodifica{\cite{Xian_2019_CVPR,Sariyildiz2019GradientMG,Gao2020ZeroVAEGANGU,GFZSL}}. Using the taxonomy of generative models  \cite{Goodfellow2017NIPS2T}, the method proposed in \cite{GFZSL} can be categorized as an explicit density model with tractable density. In fact, here it is proposed to model the density function of the conditional probability of the visual features given the class embedding through an exponential family of distributions. Among the profitable benefits of tractable density function, the computational pipeline becomes simpler and more efficient. However, constraining the density function limits the possibility to capture all the complexity of the data. Instead, our framework is based on GANs \cite{Goodfellow2014GenerativeAN,Arjovsky2017WassersteinG,improvedWGAN}, a direct implicit density model \cite{Goodfellow2017NIPS2T}, and therefore, we do not impose any density function for the distribution from which we want to generate the visual features, but we let the model to directly learn it from the data.

In \cite{Sariyildiz2019GradientMG}, a constraint is introduced in GAN training to improve the discriminative properties of the generated features. Specifically, a compatibility function $f$ between visual features and class embeddings is learned, then the correlation between gradients of real and generated features in respect to $f$ is maximized during GAN training.

In \cite{Xian_2019_CVPR} and \cite{Gao2020ZeroVAEGANGU}, a mixture of explicit and implicit models, a Variational Autoencoder (VAE) \cite{Kingma2013AutoEncodingVB} and a GAN, is proposed. Specifically, a single generator/decoder is conditioned on the attribute embeddings and used to approximate the numerically intractable distribution of the visual features. By directly minimizing the divergence between the real visual features and the generated ones, the model learns 1) how to extract visual features for those classes which are not seen during training but only described through their attributes, and 2) how to mimic the distribution of visual features (with the addition of one adversarial categorization network in \cite{Gao2020ZeroVAEGANGU}).

Differently to \cite{Xian_2019_CVPR,Sariyildiz2019GradientMG,Gao2020ZeroVAEGANGU}, in our work, we propose a decoupled feature generation framework.
Hence, instead of training one single conditional generator, we train an unconditional generator to solely capture the complexity of the visual data distribution, and we subsequently pair it with a conditional generator devoted to enrich the prior knowledge of the data distribution with the semantic content of the class embeddings.

To generalize the generative deep networks on the (unlabeled) unseen domain: \cite{Sariyildiz2019GradientMG} use  an unconditional discriminator for both, seen and unseen data, and implicitly learns the class label information through the compatibility function; \cite{Gao2020ZeroVAEGANGU} apply a pseudo-labeling strategy; \cite{Xian_2019_CVPR} use an additional unconditional discriminator for the unseen data. Differently, we cross-connect our conditional and unconditional branches.

\section{Decoupled Feature Generation}\label{sec:method}

\subsection{Notation and Problem Definition}

we consider two disjoint sets of classes: the seen  classes $\mathcal{Y}^s$ and the unseen ones $\mathcal{Y}^u$, such that $\mathcal{Y}^s\cap \mathcal{Y}^u = \emptyset $. For the seen classes, a dataset of triplets $(\textbf{x}_s,y_s,c(y_s))$ is available: $\textbf{x}_s\in \mathcal{X}$ is the visual feature vector, $y_s\in \mathcal{Y}^s$ is its class label and $c(y_s)$ is the corresponding class embedding. 
\ACCVmodifica{Differently, for the unseen classes, in transductive GZSL we only have unlabeled visual features $\textbf{x}_u$. The sets of the labels $y_u$ of the unseen classes are described in terms of their semantic embeddings $c(y_u)$, as for the seen ones.}

Given a test visual feature $\mathbf{x}$, the goal is to predict the corresponding class label $y$ which can either belong to the seen or to the unseen classes. 

For feature generation approaches, a conditional generator $G$ is fed with random noise $\mathbf{z}$, and a class embedding $c(y)$ and it synthesizes a feature vector which will be denoted by $\widetilde{\mathbf{x}}$. Once $G$ is trained, synthetic features $\widetilde{\mathbf{x}}_u$ are generated for the unseen classes and are used, together with $\mathbf{x}_s$, to train a softmax classifier which is responsible for the final recognition task.

\subsection{Our Proposed Architecture: DecGAN}\label{sec:branches}

\begin{figure}

\begin{center}
		\includegraphics[width=0.9\linewidth]{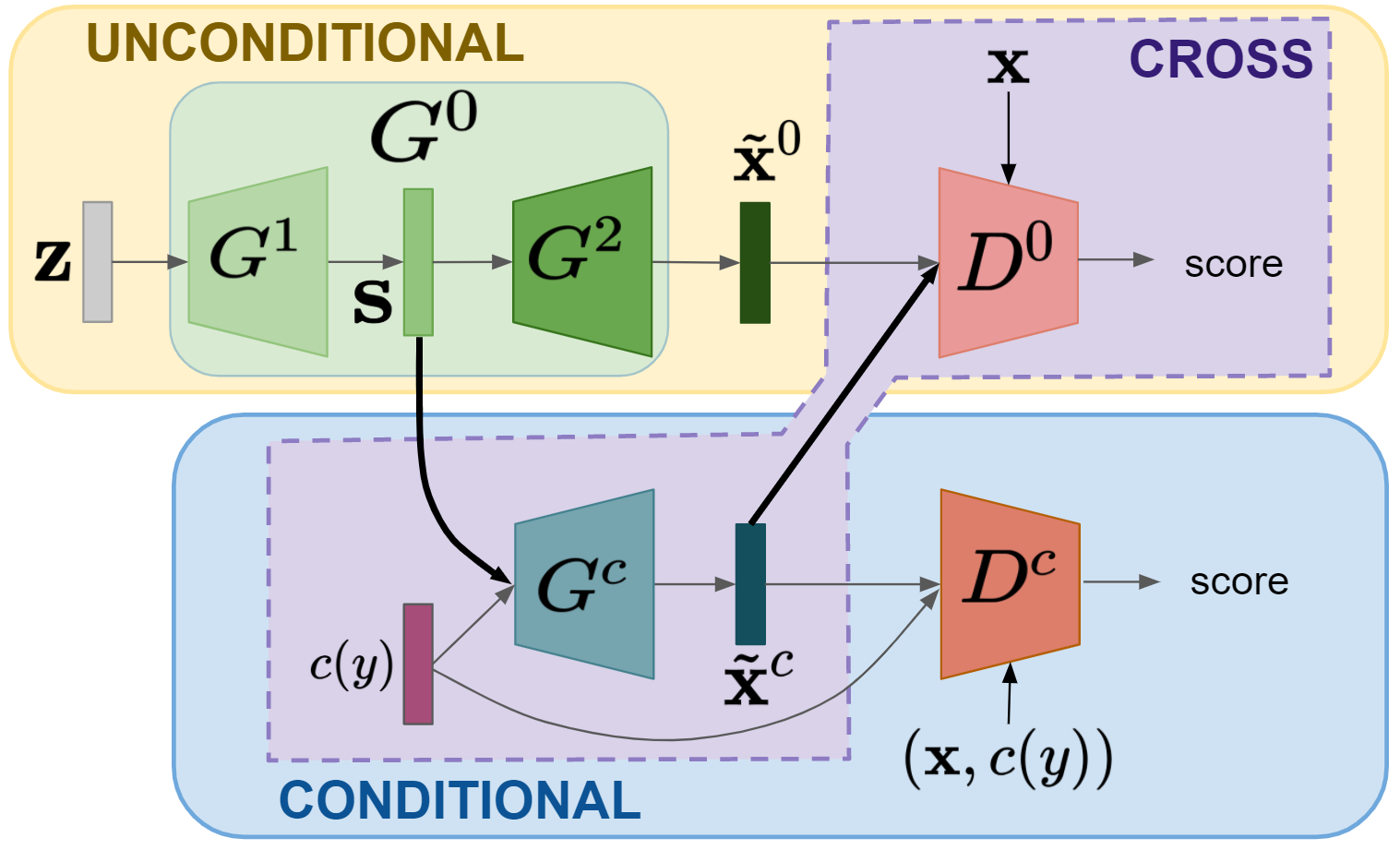}
\end{center}
		\caption{Our proposed DecGAN architecture is composed of two cross-connected branches consisting of two GANs: unconditional and conditional. The unconditional branch (yellow) is composed of generator $G^0$ and discriminator $D^0$, and the conditional branch (light blue) is composed of generator $G^c$ and discriminator $D^c$. An additional cross-branch (violet) is composed of $G^c$ and $D^0$.}\label{fig:decgan_architecture}
	Generator $G^0$ is decomposed into $G^1$ and $G^2$, such that given some random noise $\textbf{z}$, $G^2(G^1(\textbf{z}))=G^0(\textbf{z}) = \tilde{\mathbf{x}}^0$. The structured prior $\textbf{s}=G^1(\mathbf{z})$ is fed as input into $G^c$ together with the class embeddings $c(y)$, for the sake of conditionally-generating visual descriptors $\tilde{\mathbf{x}}^c$. Best viewed in colors.
\end{figure}

Looking at Figure \ref{fig:decgan_architecture}, our proposed DecGAN architecture is composed of two cross-connected branches, which consist of two GANs - one unconditional (in yellow in the figure) and one conditional (in light blue), which are cross-connected forming a third cross-branch (in violet). 

It can be noticed that there are two main ingredients: a \emph{structured prior} and a \emph{cross-connection} between the conditioned and unconditioned branches. Since the unconditional branch learns how to mimic the feature representation, regardless of the semantic class embeddings, this allows to generate a structured prior which can be shared across classes and adopted by the conditional branch to better perform the semantic-to-visual mapping.
The cross-connection is fundamental as well: once synthetic features are conditionally generated, they can be checked to be realistic from the conditional generator only if they belong to the seen classes - for which we have labels. But, with the additional usage of the unconditional generator, we can also verify if the synthetic features from unseen classes are similar to real ones in distribution. This framework fully exploits the possibility of a transductive zero-shot learning.

The reader can refer to Figure \ref{fig:decgan_architecture} for a visualization and to the next paragraphs for the details on the design of our architecture.

\subsubsection{Unconditional Branch.} The unconditional branch is composed of the generator $G^0$ and the discriminator $D^0$. The generator $G^0$ is decomposed into $G^1$ and $G^2$, such that,
given some random noise $\textbf{z}$, $G^2(G^1(\textbf{z}))=G^0(\textbf{z})$. We refer to the output of $G^1$ as the \emph{structured prior} $\textbf{s}$, that is $\textbf{s} = G^1(\textbf{z})$. The concatenation of $\textbf{s}$ and the class embeddings $c(y)$ is passed as input to the conditional branch (see next paragraph). The unconditional branch is dedicated to learning the data distribution and we model it as a Wasserstein GAN (WGAN) \cite{Arjovsky2017WassersteinG,improvedWGAN}. Hence, optimization is performed by minimizing the Wasserstein distance between the real data distribution and the synthetic's one by playing the following two-player game between $G^0$ and $D^0$ \cite{Arjovsky2017WassersteinG}:
% \vspace{-5 pt}
\begin{equation}
\min_{G^0}\max_{D^0} \mathbb{E}_{\mathbf{x}}[D^0(\mathbf{x})]- \mathbb{E}_{\tilde{\mathbf{x}}^0 }[D^0(\tilde{\mathbf{x}}^0)] ,
\label{eq:unc}
% \vspace{-5 pt}
\end{equation}
where $\tilde{\textbf{x}}^0 = G^0(\mathbf{z})$ denotes the features generated from the unconditional generator $G^0$. To regularize the min-max optimization, we use the following penalty term \cite{improvedWGAN}:
% \vspace{-5 pt}
\begin{equation}
\mathcal{R} = \mathbb{E}_{\hat{\textbf{x}}}[( \left\|\nabla D^0(\hat{\textbf{x}}) \right \|_2 -1)^2] ,
\label{eq:unc_reg}
% \vspace{-5 pt}
\end{equation}
where $\hat{\textbf{x}} = \alpha \mathbf{x} + (1 - \alpha)\tilde{\textbf{x}}^0$ with $\alpha \sim \mathcal{U}(0,1)$. 

\subsubsection{Conditional Branch.} To learn how to translate the semantic content of the class embeddings $c(y)$ we model the conditional branch with the extension of the WGAN to a conditional model \cite{Xian_2018_CVPR}. The conditional branch is composed of the generator $G^c$ and the discriminator $D^c$. In this architecture, both the generator and the discriminator are conditioned on the class embeddings. The generator $G^c$ takes as input the structured prior $\textbf{s}$ and it is conditioned on the class embeddings, learning how to enrich the information about the data distribution contained in $\textbf{s}$ with the semantic content of $c(y)$. The generated features $\tilde{\textbf{x}}^c=G^c(\textbf{s},c(y))$ are then evaluated by the discriminator $D^c$ together with the class embedding that generated them, and compared to real data pairs $(\textbf{x},c(y))$.  

With this architecture, $G^c$ learns how to enrich $\mathbf{s}$ with the content of the class embeddings. The quality of the relation between the generated visual features and the semantic content is then evaluated by $D^c$. The optimization is carried out through
% \vspace{-5 pt}
\begin{equation}
\min_{G^c}\max_{D^c} \mathbb{E}_{\mathbf{x}}[D^c(\mathbf{x},c(y))]- \mathbb{E}_{\tilde{\mathbf{x}}^c }[D^c(\tilde{\mathbf{x}}^c,c(y))] ,
\label{eq:cond}
% \vspace{-5 pt}
\end{equation}
with the regularization term \cite{Xian_2018_CVPR}:
% \vspace{-5 pt}
\begin{equation}
\mathcal{R} = \mathbb{E}_{\hat{\textbf{x}}}[( \left\|\nabla D^c(\hat{\textbf{x}},c(y)) \right \|_2 -1)^2] ,
\label{eq:cond_reg}
% \vspace{-5 pt}
\end{equation}
where $\hat{\textbf{x}} = \alpha \mathbf{x} + (1 - \alpha)\tilde{\textbf{x}}^c$ with $\alpha \sim \mathcal{U}(0,1)$. 

\WACVmodifica{We also add the regularization term introduced by \cite{felix2018multi}. That is, given a pre-trained linear module $A$ and $\tilde{a}=A(\hat{\textbf{x}})$ the reconstruction of $c(y)$ given $\hat{\textbf{x}}$, we add the reconstruction loss:
\begin{equation}
    \mathcal{R}_{rec} = \Vert c(y) - \tilde{a} \Vert_2^2.
    \label{eq:rec_loss}
\end{equation}
}

\subsubsection{Cross Branch.} Because labeled data are not available for the unseen classes, we cannot feed the conditional discriminator $D^c$ with them. To exploit the unlabeled data we propose to conditionally generate the visual features $\tilde{\textbf{x}}^c$ and evaluate them only by their distribution using $D^0$. Thus in this setting we do not condition both the generator and the discriminator, as is commonly done in GAN based conditional generation, but we only condition the  generator. Hence, optimization is obtained by 
% \vspace{-5 pt}
\begin{equation}
\min_{G^c}\max_{D^0} \mathbb{E}_{\mathbf{x}}[D^0(\mathbf{x})]- \mathbb{E}_{\tilde{\mathbf{x}}^c }[D^0(\tilde{\mathbf{x}}^c)] ,
\label{eq:trans}
% \vspace{-5 pt}
\end{equation}
adapting as consequence the regularization term on the gradients as
% \vspace{-5 pt}
\begin{equation}
\mathcal{R} = \mathbb{E}_{\hat{\textbf{x}}}[( \left\|\nabla D^0(\hat{\textbf{x}}) \right \|_2 -1)^2] ,
\label{eq:trans_reg}
% \vspace{-5 pt}
\end{equation}
where $\hat{\textbf{x}} = \alpha \mathbf{x} + (1 - \alpha)\tilde{\textbf{x}}^c$ with $\alpha \sim \mathcal{U}(0,1)$ \WACVmodifica{and adding the reconstruction loss defined in equation (\ref{eq:rec_loss})}.

\paragraph{The origin of the proposed architecture.} Our work is inspired by FusedGAN \cite{bodla2018semi}, which combines two GANs to improve image generation in a semi-supervised setup. Differently, in our case, we handle feature generation in the zero-shot case, so we have no annotated data at all for some of the classes and we need to generate them. To solve this problem, differently from FusedGAN, we cross-connect the two branches to transfer the knowledge of the seen domain to the unseen one.

\subsection{Training Methodology}\label{sec:training}

\begin{figure*}
    \centering
    \includegraphics[width=\textwidth]{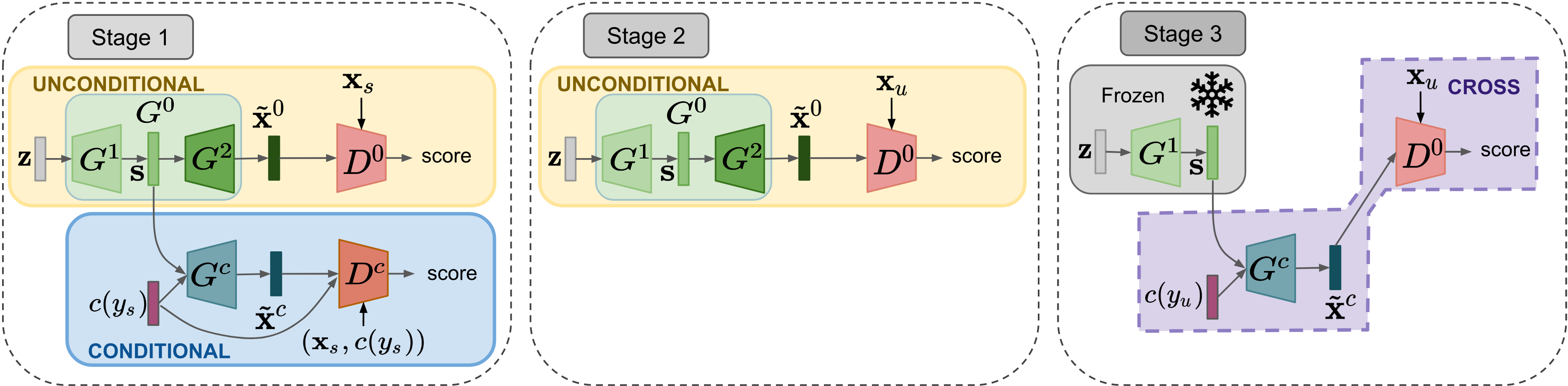}
    \caption{DecGAN training is performed in 3 stages. Stage 1 performs an alternate training on the conditional and the unconditional branch using seen data. Stage 2 uses the unconditional branch to fine-tune $G^0$ using unseen data to improve the structured prior $\mathbf{s}$. Finally, Stage 3 carries out the fine-tuning of $G^c$, feeding the cross-branch with unseen data.}
    \label{fig:decgan_training}
\end{figure*}

To train the proposed DecGAN, we propose a three-staged training strategy, which is explained beneath and sketched in Figure \ref{fig:decgan_training}.
 \begin{enumerate}
    \item In the first stage, we optimize both the conditional and the unconditional branch using only data from the seen classes.
    We seek to achieve decoupled feature generation for the seen categories in a way that, while $G^0$ learns how to model the data distribution, $G^c$ learns how to enrich the structured prior with the content of the class embeddings. We perform an alternate training strategy in which, first, we update $D^0$ and $G^0$ using equations (\ref{eq:unc}) and (\ref{eq:unc_reg}). Then, we update $D^c$ and $G^c$ using equations (\ref{eq:cond}) and (\ref{eq:cond_reg}). A full update step consists of $k$ updates of $D^0$, 1 update of $G^0$, $k$ updates of $D^c$ and 1 update of $G^c$ in sequence with $k>1$ \cite{improvedWGAN}. Here, we chose $k=5$ as done in \cite{Xian_2018_CVPR}.
    \item In the second training stage, we want to take advantage of the unseen unlabeled data to add into the structured prior the information of the unseen data distribution.
    To reach our goal, we use the unseen data to fine-tune $D^0$ an $G^0$ using equations (\ref{eq:unc}) and (\ref{eq:unc_reg}).
    \item The third stage consists in the fine-tuning of the conditional generator $G^c$ on the unseen data. Using structured prior, generalized over the unseen classes in the previous stage, we condition $G^c$ with the embeddings of the unseen classes to reinforce its ability to translate semantic content into visual features in the unseen domain. That is, we use equations (\ref{eq:trans}) and (\ref{eq:trans_reg}) to update $D^0$ an $G^c$ using unseen data.
\end{enumerate}

\subsection{Implementation Details} We implement $G^2$ and $G^c$ as single hidden layer neural networks with hidden layer of size 4096 and leaky ReLU activation and output layer that has the size of the visual feature vectors, 2048, and ReLU activation. In $G^1$, a leaky ReLU is used as the activation function (without hidden layer). Specifically, $G^0$ is a 2-hidden layer neural network and we use its first layer as structured prior. The size of the structured prior (the output of $G^1$) is fixed to $1024$. The size of noise $\textbf{z}$ is fixed to 512 and is sampled from a multivariate normal distribution $\mathcal{N}(\textbf{0},\textbf{I})$, where $\textbf{0}$ is the 512-dimensional vector of zeros and $\textbf{I}$ is the 512-dimensional identity matrix. $D^0$ and $D^c$ are neural networks composed of a single hidden layer of size 4096 (with leaky ReLU activation) and by an unconstrained real number as output.

\section{Experiments}\label{sec:exp}

\subsection{Datasets and Benchmarks}\label{sec:datasets}

We evaluate proposed DecGAN on standard benchmark datasets for GZSL. We considered Oxford Flowers (FLO) \cite{FLO}, SUN Attribute (SUN) \cite{SUN} and Caltech-UCSD-Birds 200-2011 (CUB) \cite{CUB}.
FLO consists of 8,189 images of 102 different types of flowers, SUN  14,340 images of scenes from 717 classes and CUB of 11,788 images of 200 different types of birds.

\begin{table}
\centering
\begin{tabular}{c|cccc|}
	\multicolumn{1}{c|}{Dataset} & att & stc & $\# \mathcal{Y}^s$ & $\# \mathcal{Y}^u$ \\ \hline
	FLO \cite{FLO} & -   & 1024  & 62+20        & 20             \\
	SUN  \cite{SUN} & 102 & -     & 580+65       & 72             \\ 
	CUB \cite{CUB}     & 312 & -    & 100+50       & 50             \\

	\hline
% 	\noalign{\smallskip}
\end{tabular}
\caption{Statistics of considered datasets: number of seen classes $\# \mathcal{Y}^s$ (training + validation),  unseen classes $\# \mathcal{Y}^u$, dimension of attribute (att) annotation and sentences (stc) extracted features.}
\label{tab:datsets}
\end{table}%
For SUN and CUB,
we use manually annotated attributes \cite{xian2018tPAMI}. Because for FLO the attributes are not available, we follow \cite{Xian_2019_CVPR} in using 1024-dim sentence embedding extracted by the character-based CNN-RNN  \cite{reed2016} from fine-grained visual descriptions of the images. Statistics of the datasets are available in Table \ref{tab:datsets}. For a fair comparison, we split the classes of SUN and CUB
between seen and unseen using the splits proposed by \cite{xian2018tPAMI}. For FLO, we use the splits as in \cite{reed2016}. For all datasets, the visual features are chosen as the 2048-dim top-layer pooling units of the ResNet-101 \cite{he2016deep}, provided by \cite{xian2018tPAMI}.

As evaluation metrics, the 
\ACCVmodifica{GZSL} setup, we measure the performance as harmonic mean between seen and unseen accuracy, each one computed as top-1 classification accuracy on seen and unseen classes \cite{xian2018tPAMI}.

\subsection{Ablation Study}
\label{subsec:ablation}
In this Section, we will perform an accurate ablation analysis on the different components of our proposed DecGAN architecture. Specifically, we will separately evaluate the impact on performance of each of the three branches which endows DecGAN: namely, the unconditioned, conditioned and cross-branch as presented in Section \ref{sec:branches}. We will also pay attention to the effects of the different stages of our training pipeline: the first stage is evaluated in stand-alone fashion, while we also provide experimental evidence for the effect of removing the second stage.

\subsubsection{The Impact on Performance of each Training Stage.}

To better analyse our composite training pipeline, in Table \ref{tab:ablation_training}, we present an ablation study to assess the impact on performance of each of the three stages of our training pipeline. Precisely, we evaluate the drop in performance resulting from removing any of the aforementioned stages from the full training pipeline of DecGAN: when either removing the first, second or third stage, we obtain DecGAN$^{\rm (-Stg1)}$, DecGAN$^{\rm (-Stg2)}$ and DecGAN$^{\rm (-Stg3)}$, respectively. We also assess the performance of the first and third stage separately (DecGAN$^{\rm (Stg1)}$ and DecGAN$^{\rm (Stg3)}$) Note that, we cannot evaluate the stage 2 in a standalone fashion since such stage lacks of a conditional feature generation pipeline from which we can sample features for the seen/unseen classes (see Figure \ref{fig:decgan_training}).

\begin{table*}[t!]
\centering

% \resizebox{\textwidth}{!}{
\begin{tabular}{l|ccc|ccc|ccc|}
          & \multicolumn{3}{c|}{FLO}                & \multicolumn{3}{c|}{SUN}                & \multicolumn{3}{c|}{CUB}                \\
          & $\mathbf{a}_u$    & $\mathbf{a}_s$                         & \textbf{H}    & $\mathbf{a}_u$    & $\textbf{a}_s$                         & \textbf{H}    & $\mathbf{a}_u$    & $\mathbf{a}_s$                        & \textbf{H}    \\ \hline

DecGAN$^{\rm (Stg1)}$ &  58.1 & \multicolumn{1}{c|}{79.8} & 67.2  & 45.0  & \multicolumn{1}{c|}{34.5} & 39.1 & 44.1 & \multicolumn{1}{c|}{56.7} & 49.8  \\
DecGAN$^{\rm (Stg3)}$  & 4.7  & \multicolumn{1}{c|}{82.2} & 8.9  &  1.2 & \multicolumn{1}{c|}{30.1} & 2.3  & 2.0  & \multicolumn{1}{c|}{35.3} & 3.8  \\ \hline
DecGAN$^{\rm (-Stg1)}$ &  5.8 & \multicolumn{1}{c|}{73.9} & 10.1  & 1.3  & \multicolumn{1}{c|}{39.1} & 2.5  & 1.7 & \multicolumn{1}{c|}{35.3} & 3.2 \\
DecGAN$^{\rm (-Stg2)}$   &  71.1 & \multicolumn{1}{c|}{50.5} & $80.1$ &  53.2 & \multicolumn{1}{c|}{44.2} &  $48.2$ &  55.1 & \multicolumn{1}{c|}{66.6} & $60.3$\\ 
DecGAN$^{\rm (-Stg3)}$   & 50.5 & \multicolumn{1}{c|}{80.1} & 62.0  &  44.5  & \multicolumn{1}{c|}{34.7} &  38.3 &  45.3  &  \multicolumn{1}{c|}{53.8} & 49.1\\ \hline
DecGAN   & $73.0$ & \multicolumn{1}{c|}{$92.2$} & { $\mathbf{81.5}$} & $57.2$ & \multicolumn{1}{c|}{$44.3$} & { $\mathbf{49.9}$} & $59.1$ & \multicolumn{1}{c|}{$68.4$} & $\textbf{63.4}$\\ 
\hline
%\noalign{\smallskip}
\end{tabular}%\vspace{-25pt}
% }
\caption{We assess the impact on performance of the presence/absence of each stage of the training pipeline of our DecGAN model. We report top-1 accuracy on seen classes $\mathbf{a}_s$ and unseen classes $\mathbf{a}_u$ and their harmonic mean $\mathbf{H}$. Best $\mathbf{H}$ values are highlighted in bold. All results are reported by averaging accuracies over 5 different runs. 
}
\label{tab:ablation_training}
\end{table*}%
\begin{table*}[t!]
\centering

% \vspace{5pt}
% \resizebox{0.8\textwidth}{!}{
\begin{tabular}{l|ccc|ccc|ccc|}
        %  \multirow{2}{*}{\textit{Feature Generation}} 
        & \multicolumn{3}{c|}{FLO}                & \multicolumn{3}{c|}{SUN}   & \multicolumn{3}{c|}{CUB}  \\
           & $\mathbf{a}_u$    & $\mathbf{a}_s$                         & \textbf{H}    & $\mathbf{a}_u$    & $\textbf{a}_s$                         & \textbf{H}    & $\mathbf{a}_u$    & $\mathbf{a}_s$                        & \textbf{H}    \\ \hline

Not decoupled   & 69.5 & \multicolumn{1}{c|}{ 91.4} &  79.0   & 52.7 & \multicolumn{1}{c|}{ 44.3} & 48.1 & 54.3 & \multicolumn{1}{c|}{66.7} &  59.9  \\ \hline
Decoupled   & $73.0$ & \multicolumn{1}{c|}{$92.2$} & { $\mathbf{81.5}$} & $57.2$ & \multicolumn{1}{c|}{$44.3$} & { $\mathbf{49.9}$} & $59.1$ & \multicolumn{1}{c|}{$68.4$} & $\textbf{63.4}$ \\ 
\hline
% \noalign{\smallskip}
\end{tabular}%\vspace{-15pt}
% }
\caption{The effect of decoupling the feature generation stage. We compare the decoupled approach of DecGAN to the not decoupled baseline. We report top-1 accuracy on seen classes $\mathbf{a}_s$ and unseen classes $\mathbf{a}_u$ and their harmonic mean $\mathbf{H}$. Best $\mathbf{H}$ values are highlighted in bold. All results are reported by averaging accuracies over 5 different runs.}
\label{tab:ablation_decoupling}
\end{table*}

% \subsubsection{Discussion.} 

\paragraph{Discussion.}With respect to the performance of the full DecGAN model, the first stage always achieve a suboptimal performance, and this can be clearly explained by the fact that, DecGAN$^{\rm (Stg1)}$ exploits data from the seen classes only (inductive setup). 
The poor performance of DecGAN$^{\rm (Stg3)}$ reveals that without the conditional discriminator, which takes as input the visual features and the related class embeddings (consequently learning the relation between visual features and class embeddings), the generator does not learn to generate class dependent visual features. 
When removing the first stage, the performance is comparable with DecGAN$^{\rm (Stg3)}$: this shows how pivotal this first step is for the whole pipeline.
During the second stage, we fine-tune the structured prior adding information about the data distribution of the unseen visual features, but we update the conditional generator on the new structured prior only in Stage 3. The misalignment between the structured prior that the conditional generator expects as input and the updated one it receives leads in a performance drop when comparing DecGAN$^{\rm (Stg1)}$ and DecGAN$^{\rm (-Stg3)}$. However, the importance of the transitory Stage 2, is highlighted by the difference in performance between DecGAN and DecGAN$^{\rm(-Stg2)}$. 
The third stage has a clear effect on performance: fine-tuning the conditional generator over the unseen data always boosts the performance, no matter if the second stage was performed or not. 

\subsubsection{The Effect of the Decoupled Feature Generation.}

We posit that the main advantage of DecGAN is the possibility of decoupling the feature generation stage, as to tackle two problems separately: 1) the generation of features which are visually similar to the real ones, and 2) the translation of semantic patterns from attributes to features. We want now to prove that the aforementioned separation of the tasks leads to a superior performance if compared to a model which tries to perform both tasks jointly. To do so, we compare the proposed decoupled feature generation (achieved through DecGAN) with a baseline model in which we perform a similar staged training, without performing decoupling. 

\begin{figure*}
    \centering
    \includegraphics[width=.75\textwidth]{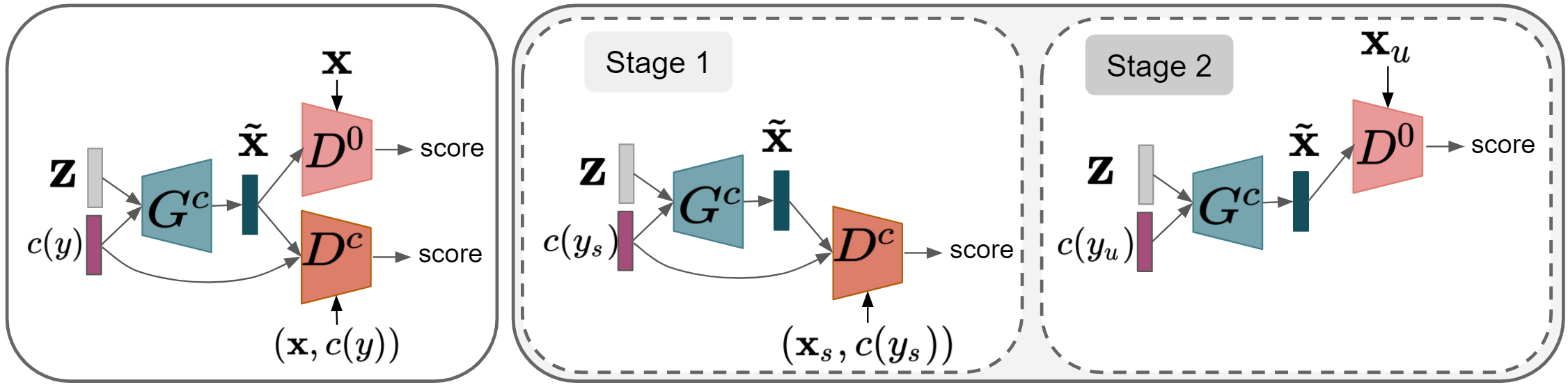}
    \caption{Baseline architecture. \emph{Left}: a visualization of the architectural design of the baseline. \emph{Right}: similarly to our proposed DecGAN, we adopt a staged training to optimize the baseline.}
    \label{fig:baselines}
\end{figure*}

Specifically, we consider the architecture represented in Figure \ref{fig:baselines}, which is described beneath. It is composed of a single conditional generator $G^c$ and two discriminators, one conditional $D^c$ and one unconditional $D^0$\WACVmodifica{, implemented through Wasserstein GANs subjects to gradient penalty loss and reconstruction loss in the same way of our DecGAN. Moreover, similarly} 
to our DecGAN, we train this architecture in two stages (see Figure \ref{fig:baselines}). In the first stage, we train $G^c$ together with $D^c$ using only seen data, and in the second stage, we train $G^c$ together with $D^0$ using unseen data. 
Differently to DecGAN, generation is performed with a single generator that has to learn both, the data distribution of the real visual features and how to translate the semantic content of the class embeddings into them, without taking advantage of the structured priors and the decoupled features generation. As for DecGAN, $G^c$ and $D^0$ and $D^c$ for this baseline are one-hidden layer neural networks with hidden layer of size 4096 with leaky ReLU activation. The size of the noise is fixed to 1024, the same of the structured prior $\mathbf{s}.$

The results of this analysis are presented in Table \ref{tab:ablation_decoupling} and discussed in the following.
\paragraph{Discussion.} The effect of decoupling is clearly visible on all the 3 benchmarks showing that it is always advantageous since leading to a superior performance, when considering all error metrics adopted in this paper: accuracy over unseen and seen classes $\mathbf{a}_u$, $\mathbf{a}_s$ and harmonic mean $\mathbf{H}$. 
When we do not perform decoupling, the resulting performance is comparable to the one achieved by DecGAN$^{\rm (-Stg2)}$ (check Tables \ref{tab:ablation_training} and \ref{tab:ablation_decoupling}). We think that this is an effect of what happens to the structure prior, which is first trained on the seen data and the conditional generator, and then fine-tuned on the unseen ones. Such discontinuous usage of seen and unseen data leads the generator to use the structured prior as a random input, since it is not able to read the visual information which is encapsulated inside.

\subsection{Comparison with the State-of-the-Art \ACCVmodifica{in transductive GZSL}}\label{sec:soa}

In this Section, we report the key benchmark against the state-of-the-art \ACCVmodifica{transductive GZSL.} 
The methods are \ACCVmodifica{ the projection with visual structure constrain (CDVSc) \cite{Wan2019TransductiveZL}, the effective deep embedding (EDE\_ex) \cite{EDE_2018}, the progressive ensamble network (PREN)  \cite{Ye2019ProgressiveEN}, the domain-invariant projection (Full DIPL) \cite{Zhao2018DomainInvariantPL},} the attribute-based latent embedding (ALE) \cite{ALE}, the generative framework based on a family of Gaussian distributions (GFZSL) \cite{GFZSL}, the discriminative semantic representation learning (DSRL) based on a non-negative matrix factorization approach \cite{Ye2017ZeroShotCW}, the feature generation approach based on a paired network and variational auto-encoder 
\ACCVmodifica{GAN and VAE \cite{Xian_2019_CVPR,Gao2020ZeroVAEGANGU} (f-VAEGAN-D2 and Z-VAE-GAN) and the addition of the gradient matching loss during the gan training \cite{Sariyildiz2019GradientMG} (GMN)}. 
\WACVmodifica{We additionally report some classical inductive methods as \cite{ESZSL, ALE, Changpinyo2016SynthesizedCF,LATEM}, as well some generative inductive methods based on GAN \cite{Xian_2018_CVPR, felix2018multi}, VAE \cite{Arora2018GeneralizedZL, Schonfeld_2019_CVPR} or combination of them \cite{Xian_2019_CVPR}.}
\WACVmodifica{We selected the three publicly available benchmark datasets (FLO, SUN and CUB) presented in Section \ref{sec:datasets}.}
We present the results through mean over five different runs at a fixed number of DecGAN training epochs.
We report results obtained in
Table \ref{tab:results_trans_ZSL_GZSL}. 
\begin{table*}[]
\centering

\begin{tabular}{l|c|ccc|ccc|ccc|}
     \multicolumn{2}{c|}{}  & \multicolumn{3}{c|}{FLO}      & \multicolumn{3}{c|}{SUN}      & \multicolumn{3}{c|}{CUB}     \\
    \multicolumn{2}{c|}{}              & $\mathbf{a}_u$  & $\mathbf{a}_s$                     & $\mathbf{H}$ & $\mathbf{a}_u$  & $\mathbf{a}_s$                     & $\mathbf{H}$ & $\mathbf{a}_u$  & $\mathbf{a}_s$                     & $\mathbf{H}$ \\ 
    \hline

\scalebox{0.8}{ESZSL} \cite{ESZSL} &  \multirow{9}{*}{I}  & -& \multicolumn{1}{c|}{-} &- & 11.0 &\multicolumn{1}{c|}{27.9} & 15.8 & 12.6 &\multicolumn{1}{c|}{63.8} & 21.0 \\
\scalebox{0.8}{ALE} \cite{ALE} &    &- & \multicolumn{1}{c|}{-} & -  & 21.8 &\multicolumn{1}{c|}{33.1} & 26.3 & 23.7 &\multicolumn{1}{c|}{62.8} & 34.4 \\
\scalebox{0.8}{SynC} \cite{Changpinyo2016SynthesizedCF} &    & -& \multicolumn{1}{c|}{-} & -& 7.9 &\multicolumn{1}{c|}{43.3} & 13.4 & 11.5 &\multicolumn{1}{c|}{70.9} & 19.8 \\
\scalebox{0.8}{LATEM} \cite{LATEM} &    &- & \multicolumn{1}{c|}{-} & - & 14.7 &\multicolumn{1}{c|}{28.8} & 19.5 & 15.2 &\multicolumn{1}{c|}{57.3} & 24.0 \\
\scalebox{0.8}{f-CLSWGAN} \cite{Xian_2018_CVPR}   &   & 59.0  & \multicolumn{1}{c|}{73.8} & 65.6  & 42.6  & \multicolumn{1}{c|}{36.6} & 39.4  &  43.7 & \multicolumn{1}{c|}{ 57.7} &  49.7 \\ 
\scalebox{0.8}{f-VAEGAN} \cite{Xian_2019_CVPR}&   &  56.8 & \multicolumn{1}{c|}{74.9} & 64.6 & 38.0 &\multicolumn{1}{c|}{45.1} & 41.3 & 48.4 &\multicolumn{1}{c|}{60.1} & 53.6 \\
\scalebox{0.8}{cycle-WGAN} \cite{felix2018multi}&  & 61.6 & \multicolumn{1}{c|}{69.2} & 65.2 & 33.8 &\multicolumn{1}{c|}{ 47.2 } & 39.4 & 47.9 &\multicolumn{1}{c|}{59.3} & 53.0 \\

\scalebox{0.8}{SyntE} \cite{Arora2018GeneralizedZL} &   & -& \multicolumn{1}{c|}{-} &- & 40.9 &\multicolumn{1}{c|}{30.5} & 34.9 & 41.5 &\multicolumn{1}{c|}{53.3} & 46.7 \\
\scalebox{0.8}{CADA-VAE} \cite{Schonfeld_2019_CVPR} &  & - & \multicolumn{1}{c|}{-} & - & 47.2 &\multicolumn{1}{c|}{35.7} & 40.6 & 51.6 &\multicolumn{1}{c|}{53.5} & 52.4 \\

    \hline
    \scalebox{0.8}{DSRL}   \cite{Ye2017ZeroShotCW} &   \multirow{11}{*}{T}  & 26.9 & \multicolumn{1}{c|}{64.3} & 37.9 & 17.7 & \multicolumn{1}{c|}{25.0} & 20.7 & 17.3 & \multicolumn{1}{c|}{39.0} & 24.0 \\ 
\scalebox{0.8}{GMN} \cite{Sariyildiz2019GradientMG} &   &  -  & \multicolumn{1}{c|}{-} & - & 57.1 & \multicolumn{1}{c|}{40.7} & 47.5 & $\dagger$  & \multicolumn{1}{c|}{$\dagger$} & $\dagger$ \\
\scalebox{0.8}{CDVSc} \cite{Wan2019TransductiveZL} &   &  -  & \multicolumn{1}{c|}{-} & - & 27.8  & \multicolumn{1}{c|}{63.2} & 38.6 & 37.0 & \multicolumn{1}{c|}{84.6} & 51.4 \\ 
\scalebox{0.8}{PREN} \cite{Ye2019ProgressiveEN} &      &  -  & \multicolumn{1}{c|}{-} & - & 35.4  & \multicolumn{1}{c|}{27.2} & 30.8 &  35.2 & \multicolumn{1}{c|}{55.8} &  43.1 \\ 
\scalebox{0.8}{Z-VAE-GAN} \cite{Gao2020ZeroVAEGANGU} &   &    - & \multicolumn{1}{c|}{-} & - &  53.1 & \multicolumn{1}{c|}{35.8 } & 42.8 & 64.1 & \multicolumn{1}{c|}{57.9} & 60.8 \\
\scalebox{0.8}{ALE\_trans} \cite{xian2018tPAMI} &   &  13.6 & \multicolumn{1}{c|}{61.4} & 22.2 & 19.9 & \multicolumn{1}{c|}{22.6} & 21.2 & 23.5 & \multicolumn{1}{c|}{45.1} & 30.9 \\ 
\scalebox{0.8}{Full DIPL} \cite{Zhao2018DomainInvariantPL} &   &  -  & \multicolumn{1}{c|}{-} & - & - & \multicolumn{1}{c|}{-} & - & 41.7 & \multicolumn{1}{c|}{44.8} & 43.2 \\ 

\scalebox{0.8}{EDE\_ex} \cite{EDE_2018}  &   &   - & \multicolumn{1}{c|}{-} & - & 47.2 & \multicolumn{1}{c|}{38.5} & 42.4 &  54.0 & \multicolumn{1}{c|}{62.9 } & 58.1 \\ 
\scalebox{0.8}{GFZSL}  \cite{GFZSL} &   &   21.8 & \multicolumn{1}{c|}{75.8} & 33.8 & 0.0  & \multicolumn{1}{c|}{41.6} & 0.0  & 24.9 & \multicolumn{1}{c|}{45.8} & 32.2\\ 
 
\scalebox{0.8}{f-VAEGAN-D2}  \cite{Xian_2019_CVPR}   &  &  78.7 & \multicolumn{1}{c|}{87.2} & \textbf{82.7} & 60.6 & \multicolumn{1}{c|}{41.9} & \textit{49.6} & 61.4 & \multicolumn{1}{c|}{65.1} & \textit{63.2}  \\ 
\cline{1-1} \cline{3-11} 
DecGAN (ours) &  &  73.0 & \multicolumn{1}{c|}{92.2} &  \textit{81.5} & 57.2 & \multicolumn{1}{c|}{44.3} & \textbf{49.9} & 59.1 & \multicolumn{1}{c|}{68.4} & \textbf{63.4}\\ \hline
\end{tabular}
% }
\label{tab:results_trans_ZSL_GZSL}
\vspace{2 pt}
\caption{
Results in GZSL. We report top-1 accuracy on seen classes $\mathbf{a}_s$ and unseen classes $\mathbf{a}_u$ and their harmonic mean $\mathbf{H}$. First and second best values are highlighted in bold and italic, respectively, for $\mathbf{H}$. Inductive (I) and Transductive (T) methods are reported. \ACCVmodifica{$\dagger$: results not reported because of the usage of different class embeddings, which are not comparable.} Our results are presented by averaging performance scores over 5 different runs.
} 
\end{table*}

\subsubsection{Discussion.} 

\WACVmodifica{In Table \ref{tab:results_trans_ZSL_GZSL}, we show how our proposed decoupled feature generation, implemented through our DecGAN model, is capable of improving in performance prior methods.}

\WACVmodifica{Among the inductive zero-shot learning methods, DecGAN sets up sharp improvements in performance: methods such as CADA-VAE \cite{Schonfeld_2019_CVPR} are improved in the scored harmonic mean $\mathbf{H}$ by +9.3\% on SUN and by +11.0\% on CUB. 
Similarly, DecGAN is capable of ouperforming cycle-WGAN \cite{felix2018multi} on FLO (+16.3\%), SUN (+10.5\%) and CUB (+10.4\%).  
Actually, even the performance of DecGAN in the first training stage is superior to cycle consistency: DecGAN$^{\rm (Stg1)}$ improves cycle-WGAN by +2.0\% of FLO.}

\WACVmodifica{A solid performance is shown also when benchmarking the state-of-the-art in the transductive generalized zero-shot learning setup. On FLO, DecGAN improves several prior methods by margin, in terms of $\mathbf{H}$: +59.3\% with respect to ALE \cite{xian2018tPAMI}, +47.7\% with respect to GFZSL \cite{GFZSL} and +43.6\% with respect to DSRL \cite{Ye2017ZeroShotCW}. The only method reported in Table \ref{tab:results_trans_ZSL_GZSL} which is slightly superior to us is f-VAEGAN-D2 \cite{Xian_2019_CVPR} and the reason for that is the usage of two feature generation schemes: a variational autoencoder and a GAN. We therefore deem that our idea is still competitive: by means of our structured prior, we can almost match the performance of a method which uses twice the number of feature generators. This evidently shows DecGAN as a balance between model light-weighting and performance. On the other two benchmark datasets, DecGAN improves f-VAEGAN-D2 \cite{Xian_2019_CVPR} by +0.3\% on SUN and +0.2\% on CUB and, similarly, surpasses in performance recent prior art such as PREN \cite{Ye2019ProgressiveEN} (+19.1\% on SUN and +20.3\% on CUB) or Z-VAE-GAN \cite{Gao2020ZeroVAEGANGU} (+7.1\% on SUN and +2.6\% on CUB).}
% \vspace{-3 pt}
\section{Conclusions}\label{sec:conc}

In this paper, we address a major limitation of the mainstream approach in (generalized) zero-shot learning, consisting in the necessity of solving two problems with a single computational pipeline: 1) capturing the distribution of visual features in order to generate realistic descriptors, and 2) translating semantic attributes into visual patterns. Therefore we proposed DecGAN, which decouples the aforementioned problems, by means of an unconditional GAN generating a structured prior. The latter can be used to improve the conditional generation of visual features. The overall architecture has a staged training, whose steps have been validated in a broad experimental comparison, assessing that this computational setup is particularly favorable for the transductive GZSL setup. In fact, DecGAN is improving in performance  previous state-of-the-art on challenging public benchmark datasets.

{\small
\bibliographystyle{ieee_fullname}
\bibliography{egbib}

\begin{thebibliography}{10}\itemsep=-1pt

\bibitem{ALE}
Zeynep Akata, Florent Perronnin, Zaid Harchaoui, and Cordelia Schmid.
\newblock Label-embedding for image classification.
\newblock {\em IEEE transactions on pattern analysis and machine intelligence},
  38(7):1425--1438, 2015.

\bibitem{Arjovsky2017WassersteinG}
Mart{\'i}n Arjovsky, Soumith Chintala, and L{\'e}on Bottou.
\newblock Wasserstein gan.
\newblock {\em ArXiv}, abs/1701.07875, 2017.

\bibitem{Arora2018GeneralizedZL}
Gundeep Arora, Vinay~Kumar Verma, Ashish Mishra, and Piyush Rai.
\newblock Generalized zero-shot learning via synthesized examples.
\newblock {\em The IEEE Conference on Computer Vision and Pattern Recognition
  (CVPR)}, 2018.

\bibitem{bodla2018semi}
Navaneeth Bodla, Gang Hua, and Rama Chellappa.
\newblock Semi-supervised fusedgan for conditional image generation.
\newblock In {\em Proceedings of the European Conference on Computer Vision
  (ECCV)}, pages 669--683, 2018.

\bibitem{Changpinyo2016SynthesizedCF}
Soravit Changpinyo, Wei-Lun Chao, Boqing Gong, and F. Sha.
\newblock Synthesized classifiers for zero-shot learning.
\newblock {\em 2016 IEEE Conference on Computer Vision and Pattern Recognition
  (CVPR)}, pages 5327--5336, 2016.

\bibitem{felix2018multi}
Rafael Felix, Vijay~BG Kumar, Ian Reid, and Gustavo Carneiro.
\newblock Multi-modal cycle-consistent generalized zero-shot learning.
\newblock In {\em The European Conference on Computer Vision (ECCV)}, 2018.

\bibitem{Fujiwara2014EfficientLP}
Yasuhiro Fujiwara and Go Irie.
\newblock Efficient label propagation.
\newblock In {\em ICML}, 2014.

\bibitem{Gao2020ZeroVAEGANGU}
Rui Gao, Xingsong Hou, Jie Qin, Jiaxin Chen, Li Liu, Fan Zhu, Zhao Zhang, and
  Ling Shao.
\newblock Zero-vae-gan: Generating unseen features for generalized and
  transductive zero-shot learning.
\newblock {\em IEEE Transactions on Image Processing}, 29:3665--3680, 2020.

\bibitem{Goodfellow2017NIPS2T}
Ian~J. Goodfellow.
\newblock Nips 2016 tutorial: Generative adversarial networks.
\newblock {\em ArXiv}, abs/1701.00160, 2017.

\bibitem{Goodfellow2014GenerativeAN}
Ian~J. Goodfellow, Jean Pouget-Abadie, Mehdi Mirza, Bing Xu, David
  Warde-Farley, Sherjil Ozair, Aaron~C. Courville, and Yoshua Bengio.
\newblock Generative adversarial nets.
\newblock In {\em NIPS}, 2014.

\bibitem{improvedWGAN}
Ishaan Gulrajani, Faruk Ahmed, Mart{\'i}n Arjovsky, Vincent Dumoulin, and
  Aaron~C. Courville.
\newblock Improved training of wasserstein gans.
\newblock In {\em NIPS}, 2017.

\bibitem{he2016deep}
Kaiming He, Xiangyu Zhang, Shaoqing Ren, and Jian Sun.
\newblock Deep residual learning for image recognition.
\newblock In {\em Proceedings of the IEEE conference on computer vision and
  pattern recognition}, pages 770--778, 2016.

\bibitem{huang2019generative}
He Huang, Changhu Wang, Philip~S Yu, and Chang-Dong Wang.
\newblock Generative dual adversarial network for generalized zero-shot
  learning.
\newblock In {\em The IEEE Conference on Computer Vision and Pattern
  Recognition (CVPR)}, 2019.

\bibitem{Kingma2013AutoEncodingVB}
Diederik~P. Kingma and Max Welling.
\newblock Auto-encoding variational bayes.
\newblock {\em CoRR}, abs/1312.6114, 2013.

\bibitem{lampert2009learning}
Christoph~H Lampert, Hannes Nickisch, and Stefan Harmeling.
\newblock Learning to detect unseen object classes by between-class attribute
  transfer.
\newblock In {\em Computer Vision and Pattern Recognition (CVPR)}. IEEE, 2009.

\bibitem{larochelle2008zero}
Hugo Larochelle, Dumitru Erhan, and Yoshua Bengio.
\newblock Zero-data learning of new tasks.
\newblock In {\em Conference on Artificial Intelligence (AAAI)}. AAAI, 2008.

\bibitem{Li:CVPR19}
Jingling Li, Mengmeng Jing, Ke Lu, Zhengming Ding, Lei Zhu, and Zi Huang.
\newblock Leveraging the invariant side of generative zero-shot learning.
\newblock In {\em The IEEE Conference on Computer Vision and Pattern
  Recognition (CVPR)}, 2019.

\bibitem{mishra2018generative}
Ashish Mishra, Shiva Krishna~Reddy, Anurag Mittal, and Hema~A Murthy.
\newblock A generative model for zero shot learning using conditional
  variational autoencoders.
\newblock In {\em The IEEE Conference on Computer Vision and Pattern
  Recognition (CVPR) Workshops}, pages 2188--2196, 2018.

\bibitem{FLO}
M-E. Nilsback and A. Zisserman.
\newblock A visual vocabulary for flower classification.
\newblock In {\em The IEEE Conference on Computer Vision and Pattern
  Recognition (CVPR)}, 2006.

\bibitem{reed2016}
Scott Reed, Zeynep Akata, Honglak Lee, and Bernt Schiele.
\newblock Learning deep representations of fine-grained visual descriptions.
\newblock pages 49--58, 06 2016.

\bibitem{ESZSL}
Bernardino Romera-Paredes and Philip Torr.
\newblock An embarrassingly simple approach to zero-shot learning.
\newblock In {\em The International Conference on Machine Learning (ICML)},
  2015.

\bibitem{salakhutdinov2011learning}
Ruslan Salakhutdinov, Antonio Torralba, and Josh Tenenbaum.
\newblock Learning to share visual appearance for multiclass object detection.
\newblock In {\em CVPR 2011}, pages 1481--1488. IEEE, 2011.

\bibitem{Sariyildiz2019GradientMG}
Mert~B{\"u}lent Sariyildiz and Ramazan~Gokberk Cinbis.
\newblock Gradient matching generative networks for zero-shot learning.
\newblock {\em 2019 IEEE/CVF Conference on Computer Vision and Pattern
  Recognition (CVPR)}, pages 2163--2173, 2019.

\bibitem{Schonfeld_2019_CVPR}
Edgar Schonfeld, Sayna Ebrahimi, Samarth Sinha, Trevor Darrell, and Zeynep
  Akata.
\newblock Generalized zero- and few-shot learning via aligned variational
  autoencoders.
\newblock In {\em The IEEE Conference on Computer Vision and Pattern
  Recognition (CVPR)}, June 2019.

\bibitem{GFZSL}
Vinay~Kumar Verma and Piyush Rai.
\newblock A simple exponential family framework for zero-shot learning.
\newblock In {\em ECML/PKDD}, 2017.

\bibitem{Wan2019TransductiveZL}
Ziyu Wan, Dongdong Chen, Yan Li, Xingguang Yan, Junge Zhang, Yizhou Yu, and
  Jing Liao.
\newblock Transductive zero-shot learning with visual structure constraint.
\newblock In {\em Advances in Neural Information Processing Systems 32}, pages
  9972--9982. 2019.

\bibitem{CUB}
P. Welinder, S. Branson, T. Mita, C. Wah, F. Schroff, S. Belongie, and P.
  Perona.
\newblock {Caltech-UCSD Birds 200}.
\newblock Technical Report CNS-TR-2010-001, California Institute of Technology,
  2010.

\bibitem{LATEM}
Yongqin Xian, Zeynep Akata, Gaurav Sharma, Quynh Nguyen, Matthias Hein, and
  Bernt Schiele.
\newblock Latent embeddings for zero-shot classification.
\newblock In {\em Proceedings of the IEEE Conference on Computer Vision and
  Pattern Recognition}, pages 69--77, 2016.

\bibitem{xian2018tPAMI}
Yongqin Xian, Christoph~H Lampert, Bernt Schiele, and Zeynep Akata.
\newblock Zero-shot learning: a comprehensive evaluation of the good, the bad
  and the ugly.
\newblock {\em The IEEE Transactions on Pattern Analysis and Machine
  Intelligence}, 2018.

\bibitem{Xian_2018_CVPR}
Yongqin Xian, Tobias Lorenz, Bernt Schiele, and Zeynep Akata.
\newblock Feature generating networks for zero-shot learning.
\newblock In {\em The IEEE Conference on Computer Vision and Pattern
  Recognition (CVPR)}, June 2018.

\bibitem{Xian_2019_CVPR}
Yongqin Xian, Saurabh Sharma, Bernt Schiele, and Zeynep Akata.
\newblock {F-VAEGAN-D2: A Feature Generating Framework for Any-Shot Learning}.
\newblock In {\em The IEEE Conference on Computer Vision and Pattern
  Recognition (CVPR)}, June 2019.

\bibitem{SUN}
J. {Xiao}, J. {Hays}, K.~A. {Ehinger}, A. {Oliva}, and A. {Torralba}.
\newblock Sun database: Large-scale scene recognition from abbey to zoo.
\newblock In {\em The IEEE Conference on Computer Vision and Pattern
  Recognition (CVPR)}, 2010.

\bibitem{Ye2017ZeroShotCW}
Meng Ye and Yuhong Guo.
\newblock Zero-shot classification with discriminative semantic representation
  learning.
\newblock {\em 2017 IEEE Conference on Computer Vision and Pattern Recognition
  (CVPR)}, pages 5103--5111, 2017.

\bibitem{Ye2019ProgressiveEN}
Meng Ye and Yuhong Guo.
\newblock Progressive ensemble networks for zero-shot recognition.
\newblock {\em 2019 IEEE/CVF Conference on Computer Vision and Pattern
  Recognition (CVPR)}, pages 11720--11728, 2019.

\bibitem{EDE_2018}
Lei Zhang, Peng Wang, Lingqiao Liu, Chunhua Shen, Wei Wei, Yanning Zhang, and
  Anton van~den Hengel.
\newblock Towards effective deep embedding for zero-shot learning.
\newblock {\em ArXiv}, abs/1808.10075, 2018.

\bibitem{Zhao2018DomainInvariantPL}
An Zhao, Mingyu Ding, Jiechao Guan, Zhiwu Lu, Tao Xiang, and Ji-Rong Wen.
\newblock Domain-invariant projection learning for zero-shot recognition.
\newblock In {\em NeurIPS}, 2018.

\bibitem{Zhu:ICCV19}
Yizhe Zhu, Jianwen Xie, Bingchen Liu, and Ahmed Elgammal.
\newblock Learning feature-to-feature translator by alternating
  back-propagation for generative zero-shot learning.
\newblock In {\em The IEEE International Conference on Computer Vision (ICCV)},
  2019.

\end{thebibliography}
\balance
}

\end{document}